\documentclass[letterpaper]{article} 
\usepackage{aaai25}  
\usepackage{times}  
\usepackage{helvet}  
\usepackage{courier}  
\usepackage[hyphens]{url}  
\usepackage{graphicx} 
\usepackage{natbib}  
\usepackage{caption} 
\usepackage{algorithm}

\usepackage{xspace}
\usepackage{xcolor}
\usepackage{multirow}
\usepackage{enumitem}
\usepackage{nicefrac}       
\usepackage{amsfonts}       
\usepackage{booktabs}
\usepackage{stmaryrd}
\usepackage{placeins} 
\usepackage{amsmath} 
\usepackage{algorithmicx}
\usepackage{algpseudocode}
\usepackage{newfloat}
\usepackage{listings}
\DeclareCaptionStyle{ruled}{labelfont=normalfont,labelsep=colon,strut=off} 
\floatstyle{ruled}
\newfloat{listing}{tb}{lst}{}
\floatname{listing}{Listing}
\title{ReX: A Framework for Incorporating Temporal Information \\in Local Model-Agnostic Explanation Techniques}
\author {
    Junhao Liu,
    Xin Zhang\thanks{Corresponding author.}
}
\affiliations {
    Key Lab of High Confidence Software Technologies (Peking University), Ministry of Education, Beijing, China\\
    School of Computer Science, Peking University, Beijing, China\\
    liujunhao@pku.edu.cn, xin@pku.edu.cn
}

\begin{document}

\def\toolname{\textsc{ReX}\xspace}
\def\tobecheck{\textcolor{red}{\{to be checked\}}}
\def\ran{\mathrm{ran}}
\def\neigh{\mathrm{neigh}}
\def\per{\mathrm{per}}

\newtheorem{theorem}{Theorem}
\newtheorem{proposition}[theorem]{Proposition}
\newtheorem{lemma}[theorem]{Lemma}
\newtheorem{corollary}[theorem]{Corollary}
\newtheorem{definition}[theorem]{Definition}
\newtheorem{assumption}[theorem]{Assumption}
\newtheorem{remark}[theorem]{Remark}

\maketitle

\begin{abstract}
Existing local model-agnostic 
explanation techniques are ineffective for machine learning models that consider inputs of variable lengths, as they do not consider temporal information embedded in these models. 
To address this limitation, we propose \textsc{ReX}, a general framework for incorporating temporal information in these techniques.
Our key insight is that these techniques typically learn a model surrogate by sampling model inputs and outputs, and we can incorporate temporal information in a uniform way by only changing the sampling process and the surrogate features. 
We instantiate our approach on three popular explanation techniques: Anchors, LIME, and Kernel SHAP. 
To evaluate the effectiveness of \textsc{ReX}, we apply our approach to six models in three different tasks.
Our evaluation results demonstrate that our approach 1) significantly improves the fidelity of explanations, making model-agnostic techniques outperform a state-of-the-art model-specific technique on its target model, and 2) helps end users better understand the models' behaviors.
\end{abstract} 

\begin{links}
    \link{Extended version}{https://arxiv.org/abs/2209.03798}
\end{links}

\section{Introduction}

\label{sec:intro}

As more critical applications employ machine learning models, how to explain the rationales behind these models has emerged as an important problem.
Such explanations allow end users to 1) judge whether the results are trustworthy~\cite{LIME, account} and 2) understand knowledge embedded in the models, so they can use the knowledge to manipulate future events~\cite{PoyiadziSSBF20, ProsperiGSKMHRW20, ZhangSS18}.
This paper focuses on the problem of explaining deep models
processing sequential data of variable lengths, such as Recurrent Neural Networks (RNNs) and Transformers~\cite{vaswani2017attention,wolf-etal-2020-transformers} including large language models (LLMs)~\cite{Llama2,GPT4}.

To faithfully describe the behaviors of these models, it is important to consider the effect of temporal information,
as the models care about not only the values of features but also their positions when making decisions.
Unfortunately, existing techniques either consider temporal information but fail to produce faithful explanations that are understandable, or do not consider it at all and therefore produce explanations of low fidelity.
%
%
%
%
All existing techniques that consider temporal information are global~\cite{Jacobsson05, WangZOXLG18} (e.g. surrogate deterministic finite automatons~\cite{OmlinG96, WeissGY18, DongWSZWDDW20}), which explain target models on the whole input domain~\cite{survey_XAI}. 
%
However, faithful global explanations are complex for real-world models, which renders them hard for end users to understand, and limits their application in practice.
In contrast, local techniques~\cite{LIME, Anchors, ZhangSS18, ArrasMMS17, Counterfactual, SHAP} explain target models on a particular set of inputs (typically ones that are similar to a given input), so they can produce more tractable and understandable explanations~\cite{survey_zhang}.
However, none of them captures the effect of temporal information, which leads to low fidelity.
%


\begin{figure}
    \centering
  \includegraphics[scale=0.49]{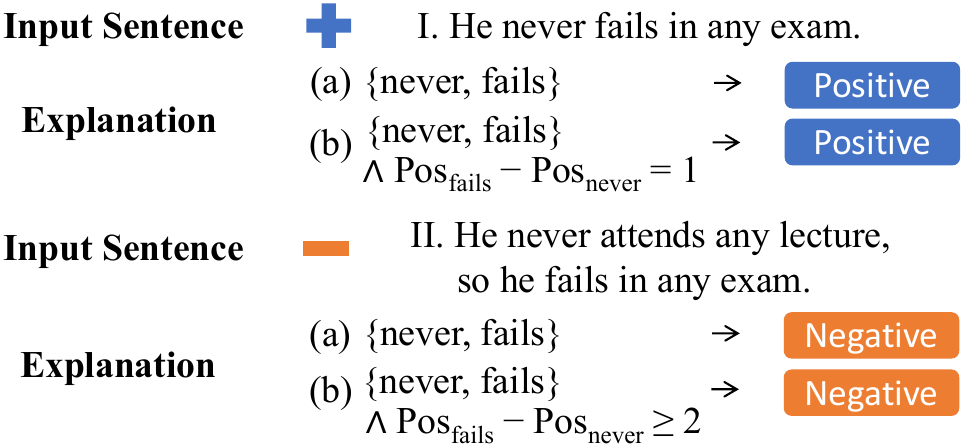} 
    \caption{Example Anchors explanations (a) and REX-augmented Anchors explanations (b).}
    \label{fig:example1new}
\end{figure}

\begin{figure}
    \centering
    \small
    \begin{tabular}{@{}l@{\ }l}
        \textbf{Input sentence}:& Bob is not a bad boy. \\
        \textbf{Network output}: & Positive  \\
             \textbf{Explanations}:&\\

    \end{tabular}
    \begin{tabular}{r@{}l}
  \includegraphics[scale=0.365]{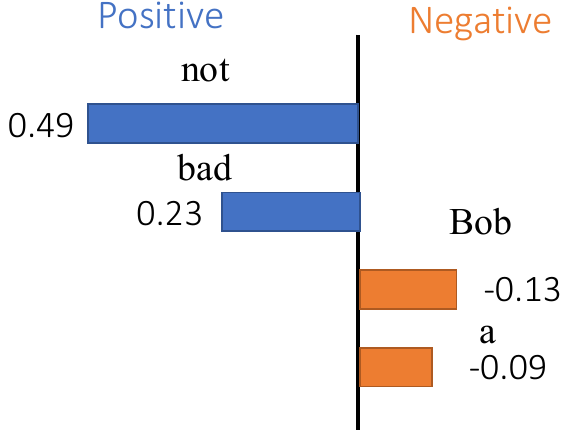}  &
  \includegraphics[scale=0.365]{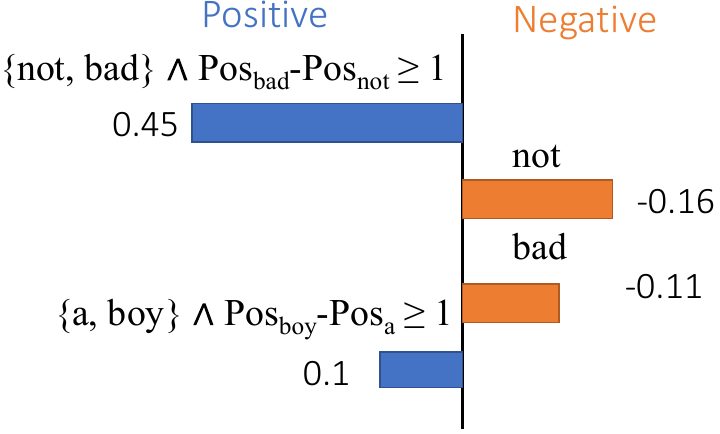}
    \end{tabular}
    \caption{Example explanations generated by LIME (left) and \toolname-augmented LIME (right).}
    \label{fig:example2}
\end{figure}

To bridge this gap,
we plan to incorporate temporal information into various popular local explanations.
Moreover, to ensure our method can explain a wide range of models, we focus on local model-agnostic techniques.  
Towards this end, we propose \toolname, a general framework for incorporating temporal information in various local model-agnostic explanation techniques.

We take two popular local model-agnostic explanation techniques, Anchors~\cite{Anchors} and LIME~\cite{LIME}, as examples to show how our framework improves existing techniques.

Figure~\ref{fig:example1new} and~\ref{fig:example2} show explanations for two sentiment analysis models.
Figure~\ref{fig:example1new} shows the Anchors explanations (referred to as \emph{anchors}) of an LSTM on two sentences.
Anchors provides rule-based local sufficient conditions for model predictions.
For Sentence I, the anchor states that 
\emph{the presence of ``never'' and ``fails'' guarantees a positive prediction}. 
For Sentence II, the anchor is the same, but the prediction is negative.
The key difference is that the words ``never'' and ``fails'' form a phrase in Sentence I, whereas they are separated in Sentence II.
The anchors fail to capture the difference, and their infidelity leads to confusing results.
Explanations with temporal information address this issue.
\toolname-augmented explanations use $Pos_{w}$ to denote the position of a word $w$ in the sentence, i.e., $w$ is the $Pos_{w}$-th word in the sequence.
For Sentence I, the \toolname-augmented anchor states that \emph{the presence of ``never'' and ``fails'' with ``never" right before ``fails" guarantees a positive prediction}. 
For Sentence II, the \toolname-augmented explanation is \emph{the presence of ``never'' and ``fails'' with ``never" \textbf{not} right before ``fails" guarantees a negative prediction}.
The \toolname-augmented anchors faithfully capture the behaviors of the LSTM.

Similar issues exist in LIME, which provides feature attributions.
%
Figure~\ref{fig:example2} shows a LIME explanation of a BERT-based sentiment analysis model~\cite{devlin2018bert} on a sentence.
LIME assigns high positive scores to the words ``not" and ``bad".
It indicates that either ``not" or ``bad" can make the sentence more positive, which is unfaithful to the BERT model.
In this way, users can predict ``Bob is a bad boy" as a positive sentence, as the word ``bad" should have a strong positive effect.
However, the model prediction is negative.
The key is that ``not bad" together is a positive phrase, whereas ``not" or ``bad" alone is a negative word.
Incorporating temporal information in LIME addresses this issue.
\toolname-augmented LIME gives \emph{``not" before ``bad"} the highest positive score, whereas \emph{``not"} and \emph{``bad"} both get negative scores, 
which 1) associates the two words, and 2) captures that ``not" comes before ``bad."

\begin{figure}[t]
    \centering
    \includegraphics[scale=0.28]{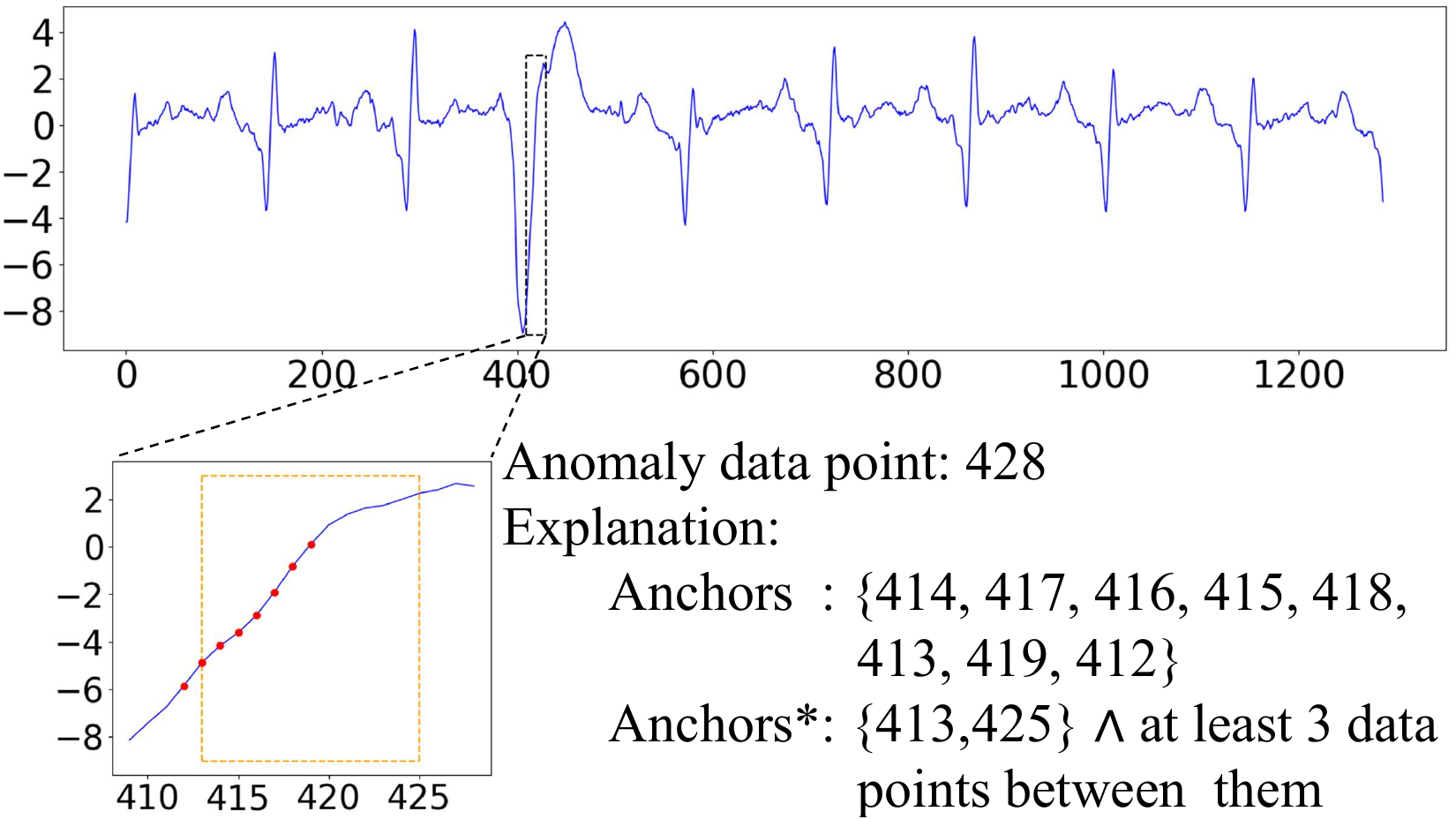}
    \caption{Anchors and \toolname-augmented Anchors~(Denoted as Anchors*) explanations for an anomaly detection RNN. Anchors: if the values of $x_{414}$, $x_{417}$, $x_{416}$, $x_{415}$, $x_{418}$, $x_{413}$, $x_{419}$, and $x_{412}$ remain unchanged, $x_{428}$ will be classified as an anomaly. Anchors*: if the values of $x_{413}$ and $x_{425}$ remain unchanged, and there are at least 3 data points between them, $x_{428}$ will be classified as an anomaly.}
    \label{fig:adexp}
\end{figure}

Figure~\ref{fig:adexp} shows another example of explaining an anomaly detection RNN that takes a time series data $x = x_1x_2...x_n$ as input. After reading $x_1x_2...x_i$, the RNN outputs a binary label $y_i$ to indicate whether $x_i$ is an anomaly.
An anchor is that \emph{the anomaly is detected because of the presence of several separated data points}.
The \toolname-augmented anchor states that \emph{the anomaly is detected because of the presence of data points 413 and 425 with at least 3 points between them.}
%
The \toolname-augmented explanation is more faithful and reveals more meaningful information to end users.

The preceding examples indicate that existing local explanations can be unfaithful and confusing for model processing inputs of variable lengths, 
because existing techniques only use the values of {\bf features} (e.g. words for text data, data points for time-series data) as the components to build explanations.
%
%
%
To address this issue, 
\toolname adds temporal information by showing the effect of the absolute positions of features and the relative positions between features.
The examples show that incorporating temporal information improves the fidelity, making users better understand the behaviors of the models. 
%


To incorporate temporal information into various local model-agnostic explanation techniques in a uniform and lightweight manner, 
we have made two key observations: 1) these techniques use a perturbation model to generate samples that are similar to the original input, and capture the local behavior of the model via these samples; 2) these techniques use feature predicates as the basic language components to construct explanations.
Therefore, by only extending the perturbation model and the language components of explanations, \toolname enables these techniques to incorporate temporal information automatically without changing their core algorithms.

We demonstrate the effectiveness of \toolname by evaluating the explanations of an LSTM, four transformer models (BERT, T5~\cite{T5}, GPT-2~\cite{GPT2}, Llama 2~\cite{Llama2}) on a sentiment analysis task, an RNN on an anomaly detection task, and Llama 2 on a text generation task, 
after applying \toolname to Anchors, LIME, and Kernel SHAP~\cite{SHAP}.
On average, \toolname helps improve the fidelity of Anchors, LIME, and Kernel SHAP explanations by 218.7\%, 41.2\%, and {36.0\%} respectively. Moreover, \toolname-augmented LIME and SHAP outperform DecompX~\cite{DecompX}, a state-of-the-art explanation method for text models on its target model.
We also run a user study, which shows that \toolname helps end users better understand and predict the behaviors of target models.

\section{Preliminaries}
\label{sec:pre}

\begin{figure*}[t]
    \centering
     \includegraphics[scale = 0.38]{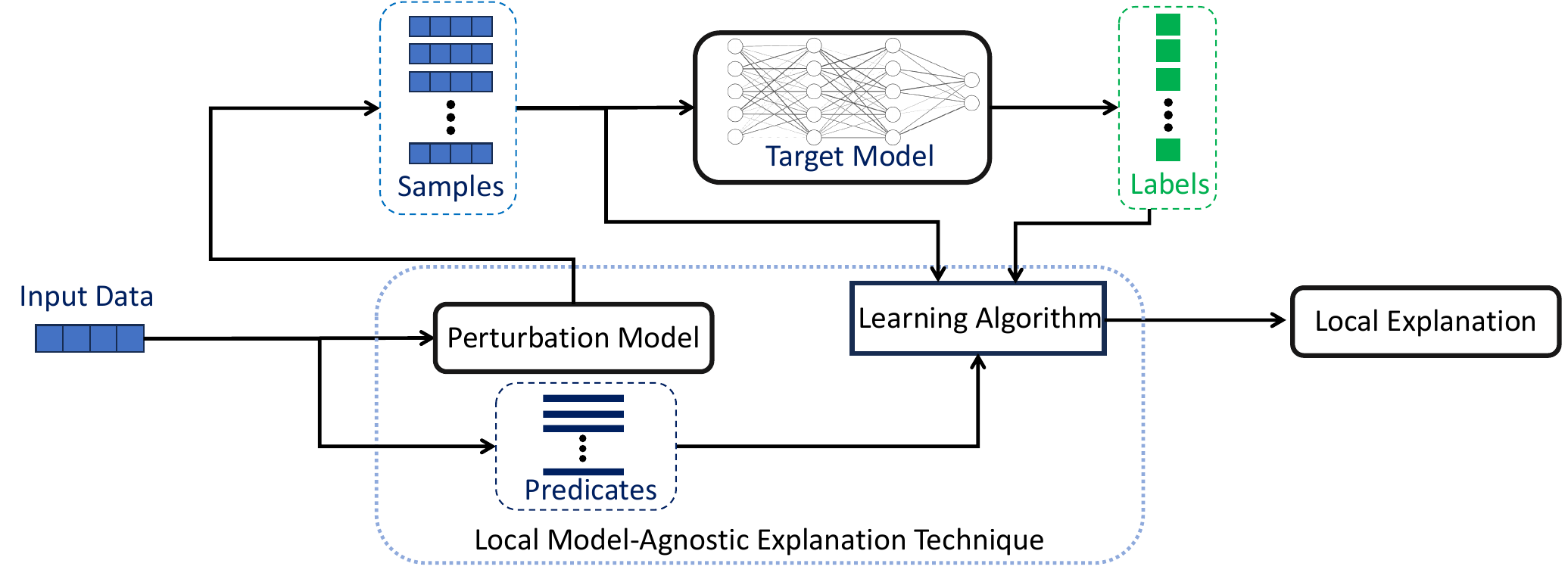}
    \caption{The workflow of generating explanations by a local model-agnostic explanation technique.}
    \label{fig:pipe}
\end{figure*}

In this section, we introduce the background of our approach.
Without loss of generality, we assume the target model is a black-box function from a sequence of real numbers to a real number, $f: \mathbb{R}^{*} \rightarrow \mathbb{R}$, where $\mathbb{R}^*=\bigcup_{T \in \mathbb{N}} \mathbb{R}^T$. For an input $x=(x_1,x_2,\dots, x_n)$, let $|x|$ denote the length of $x$, and $x_i$ represent the $i$-th feature value of $x$.
We limit our discussion to classifiers and regressors.

Given an input $x\in \mathbb{R}^{*}$ and a model $f$, a local model-agnostic explanation technique $t$ generates a local explanation $g_{f,x} := t(f, x)$. This explanation, denoted as $g$ for simplicity, is a self-interpretable expression that describes the behaviors of $f$ around $x$ formed with \textbf{predicates}.
Each predicate $p$ maps an input to a binary value, i.e., $p: \mathbb{R}^* \rightarrow \{0, 1\}$.

All existing local model-agnostic explanation techniques generate explanations in a similar workflow, as shown in Figure~\ref{fig:pipe}, which involves three steps: 

\begin{enumerate}[itemsep=0pt,parsep=0pt,topsep=0 pt,partopsep=0pt,leftmargin=0.5cm]
    \item \textbf{Producing  Predicates}: These techniques produce a set of predicates based on the input $x$, denoted as $P$.
    \item \textbf{Generating Samples}: An underlying perturbation model $t_{per}$ generates a set of samples that are similar to the input $x$, denoted as $X_s$.
    \item \textbf{Learning Explanations}:  These techniques learn a local explanation $g$ consisting of predicates in $P$, using $X_s$ and its corresponding model outputs $f(X_s)$.
\end{enumerate} 

Popular techniques such as LIME, Anchors, and Kernel SHAP all follow this workflow. Specifically, they use the same kinds of predicate sets and perturbation models. In the following, we introduce their predicate sets, perturbation models, and learning algorithms in detail.

\textbf{Predicate Sets.} 
Given an input $x$, and letting $\llbracket 1, n \rrbracket := \{1,2,\dots,n\}$, a predicate set is defined as follows:
\begin{equation}
    P:= \{p_{i} | i \in \llbracket 1, |x| \rrbracket \}.
    \label{eq:p}
\end{equation}
Here, each $p_{i}$ is a \textbf{feature predicate} defined by \(p_{i}(z) := 1_{\ran(x,i)}(z_i)\), where $\ran(x, i)$ is a set containing $x_i$.
Specifically, $p_i$ is an indicator function that checks if the $i$-th feature of a sample $z$ has a similar value to $x_i$ (i.e., $z_i\in \ran(x, i)$). For example, we can use $\ran(x,i) = (x_i-\epsilon,x_i+\epsilon)$ for a real number $x_i$, and use $\ran(x,i) = \{x_i\}$ for a categorical value $x_i$.

\textbf{Perturbation Models.} 
The perturbation model $t_{per}$ generates samples by changing the feature values of the input $x$. Given a parameter $n$, $t_{per}$ generates $n$ samples from the domain $D$ defined as follows:
\begin{equation}
D=\{z\in \mathbb{R}^{|x|}\mid \forall i \in \llbracket 1, |x| \rrbracket, z_i \in \per(x, i)\}
\label{eq:per}
\end{equation}
where $\per(x, i)$ is the perturbation range of $x_i$, whose definition depends on the type of $x_i$. Specifically, $\per(x, i)$ contains values similar to $x_i$ and \(\ran(x,i)\subset\per(x,i)\).

%

\textbf{Learning Algorithms and Explanations.} 
This step is to learn an understandable expression $g$.
In Anchors, $g$ is a conjunction of predicates that provides a sufficient condition for $f$ to produce the same output as $f(x)$, i.e., $f(z) = f(x)\ \text{if}\ g(z) = 1$. 
Specifically, $g(z) = \bigwedge_{p\in Q} p(z)$, where  $Q$ is selected from $P$ by a greedy algorithm based on the \text{KL-LUCB} algorithm~\cite{KL-LUCB}. 
In LIME and kernel SHAP, $g$ is a linear expression that serves as a local surrogate model of $f$, i.e.,
$g(z) = \Sigma_{p\in P}\ \omega_p p(z) + \omega_0$, 
where $\omega_p$ is the weight of $p$ and $\omega_0$ is a constant. 
LIME and kernel SHAP use different linear regression algorithms to learn $\omega_p$.

Due to the limitations of the preceding predicates and perturbation models, existing local explanation techniques can
only capture the behavior of target models on samples of the same lengths as the original input\footnote{When explaining NLP models, some perturbation 
models allow replacing a word with an empty string. This enables the explanations to cover inputs of shorter lengths to some extent.}
, and produce explanations with only constraints of feature values,
which limits their effectiveness on models processing sequential data of different lengths.
\section{The \toolname (tempoRal eXplanation) Framework}

\label{sec:frame}

%
We propose \toolname to provide a general approach to incorporate temporal information in explanations, without requiring significantly modifying existing techniques.
In this section, we introduce \toolname in three steps: 
1) defining local explanations with temporal information, 
2) showing how to augment existing techniques to generate these explanations, 
and 3) outlining the \toolname-augmented workflow.

\subsection{Local Explanations with Temporal Information}
Our key observation is that although the form of explanation expressions varies, the expressions are all built from the predicate set ${P}$.
If we can use predicates that reflect temporal information to build explanations, temporal information is inherent in the explanations.

Our temporal predicates describe the temporal relationship between a set of features.
We limit the number of features in a temporal predicate up to two because 1) in most cases, the temporal relationship between two features suffices to cover a large range of inputs of different lengths, and 2) humans have difficulty understanding high-dimensional information.
%
Their definitions are as below:

\begin{definition}[1-D Temporal Predicate] Given an input $x$, a 1-D temporal predicate takes the form of 
\label{def:1d}
\begin{equation}
p_{k,d,op}^{1D}(z) := \exists i \in \mathbb{Z}^{+},\ (z_i\in \ran(x,k) \wedge i\ op\ d)
\end{equation}
where $d\in \mathbb{Z}^{+}$, and $op$ is a binary operator, like $=,\le$, and $\ge$. 
\end{definition}

\begin{definition}[2-D Temporal Predicate] Given an input $x$, a 2-D temporal predicate takes the form of
\begin{multline}
p_{k,l,d,op}^{2D}(z) :=
    \exists i, j \in \mathbb{Z}^{+},\  \\(
    z_i\in \ran(x,k) \wedge z_j \in \ran(x,l) \wedge j-i\ op\ d)
\end{multline}
where $d\in \mathbb{Z}$, and $op$ is a binary operator.

\label{def:2d}
\end{definition}

%

We use 1-D temporal predicates to illustrate the effect of a single feature’s absolute position, and 2-D to illustrate the effect of the relative position between two features.
Moreover, 2-D temporal predicates also apply to the case where the presence of two features together is important, but their order is not.
We then introduce the definition of local explanations with temporal information:
\begin{definition} [Explanation with Temporal Information] 
A local explanation with temporal information is a local explanation constructed from feature predicates, 1-D temporal predicates, and 2-D temporal predicates.
\end{definition}

\textbf{Examples.} Given Input Sentence I in Figure~\ref{fig:example1new}, 
a 2-D temporal predicate is
\(
    \exists i,j \in \mathbb{Z}^{+},\ (
    z_i = ``never" \wedge z_j = ``fails" \wedge j - i = 1).
\)
The \toolname-augmented anchor in Figure~\ref{fig:example1new} is a conjunction consisting of only the preceding 2-D predicate.
For another sentence ``He could solve only the problem," which is judged as negative, the \toolname-augmented anchor is
\(
\exists i\in \mathbb{Z}^+,\ (z_i=``only" \wedge i \ge 2).
\)
This 1-D predicate indicates that similar sentences with ``only" not at the beginning are judged as negative, such as ``She only could solve the problem." However, sentences like ``Only he could solve the problem" are neutral or positive.
%

\subsection{Augmenting Generation Techniques}
Consider the workflow of existing local model-agnostic techniques in Figure~\ref{fig:pipe}. 
While their core algorithms differ, they are not different from standard machine learning algorithms at a high level.
To make them incorporate temporal information without modifying their underlying design, we only need to change their features and the data they learn from:
To incorporate the temporal predicates into explanations, we need to extend the predicate set $P$; to capture the effect of temporal information, we need to extend the perturbation model $t_{per}$.

\textbf{Extending Predicate Sets.} 
Definition~\ref{def:1d} and~\ref{def:2d} provide the forms of 1-D and 2-D temporal predicates.
Given an input $x$, we add the corresponding 1-D temporal predicates
\begin{equation}   
        P_{1D}:=  \{p_{k,d,op}^{1D}| k,d\in \llbracket 1, |x| \rrbracket \wedge
        op \in \{=, >, <,\geq,\leq\}\}
    \label{eq:1d}
\end{equation}
and 2-D temporal predicates
\begin{multline}   
        P_{2D} := \{p_{k,l,d,op}^{2D}|k,l\in \llbracket 1, |x| \rrbracket \wedge k<l \\
       \wedge d\in \llbracket -|x|, |x| \rrbracket \wedge op \in \{=, >, <,\geq,\leq\}\}
    \label{eq:2d}
\end{multline}
to the predicate set. 
%
For a specific usage scenario, users can further restrict the range of $k$,$l$,$d$, and $op$. 
For example, 
users can set a window $w$ to limit $|d-k|\le w$ for $P_{1D}$, and $|d|\le w$ and $l-k\le w$ for $P_{2D}$.
We define the \toolname extended predicate set as 
\(P^R := P \cup P_{1D} \cup P_{2D}\).


\textbf{Extending Perturbation Models.} To generate samples of different lengths, 
%
we add a postprocessor to the perturbation models of existing techniques.
The postprocessor can generate samples of different lengths by removing or switching features.
For an input $x$, the postprocessor does feature removal and swap on it in sequence: 1) $rf(x)$ returns a set of all subsequences of $x$; 2) $sf(x,i,j)$ switches the i-th and j-th features of $x$. 
%
%
We denote the \toolname-augmented perturbation model as $t_{per}^R$, and $t_{per}^R(x,n)$ takes $n$ samples from the domain $D^R$ defined as follows:
\begin{multline}
D^R = \{ \text{sf}(z, i, j) \mid z \in \bigcup_{z' \in t_{\text{per}}(x)} \text{rf}(z') \\
\wedge i, j \in \llbracket 1, |z| \rrbracket \wedge i < j\}.
\end{multline}
%
%
%
%
%

\subsection{\toolname-Augmented Workflow}
Compared to the vanilla workflows shown in Figure~\ref{fig:pipe}, the \toolname-augmented techniques use similar workflows, but replace the predicate set $P$ with $P^R$ and the perturbation model $t_{per}$ with $t_{per}^R$.
%
%
As a result, they can capture target models' behaviors on variable-length inputs by $t_{per}^R$-generated samples, and present the effect of temporal information with temporal predicates in $P^R$.


\section{Empirical Evaluation}
\label{sec:eval}

\begin{table*}[t]
    \centering
    {
    \setlength{\tabcolsep}{1mm}
    \renewcommand{\arraystretch}{0.9}
    \small
    \begin{tabular}{@{}lcccccccccccc@{}}
        \toprule
        \multirow{2}{*}{Method} & \multicolumn{6}{c}{Precision (\%)} & \multicolumn{6}{c}{Coverage (\%)} \\
        \cmidrule(lr){2-7} \cmidrule(l){8-13}
       & LSTM & BERT & T5 & GPT-2 & Llama 2 & Anom. & LSTM & BERT & T5 & GPT-2 & Llama 2 & Anom. \\
       \midrule
       Anchor & 84.74 & 82.64 & 81.74 & 82.03 & 79.15 & \textbf{90.40} &1.87 & 2.46 & 1.29 & 8.08 & 1.75 & 4.60 \\
       Anchor* &  \textbf{86.48} & \textbf{84.04} & \textbf{84.04} & \textbf{82.54} & \textbf{80.20} & 89.10 & \textbf{10.77} & \textbf{12.07} & \textbf{11.78} & \textbf{10.26} & \textbf{10.35} & \textbf{8.70}\\
       \bottomrule
    \end{tabular}
    }
    \caption{Average precision and coverage of anchors and ReX-augmented anchors~(denoted as anchors*) for sentiment analysis models and the anomaly detection RNN~(Anom.).}.
    \label{table:fidelity-anchor}
\end{table*}

\begin{table*}[t]
    \centering
    {
    \setlength{\tabcolsep}{1mm}
    \renewcommand{\arraystretch}{0.9}
    \small
   \begin{tabular}{@{}lcccccccccccc@{}}
        \toprule
        \multirow{2}{*}{Method} & \multicolumn{6}{c}{Accuracy (\%)} & \multicolumn{6}{c}{AUROC}\\
        \cmidrule(lr){2-7} \cmidrule(l){8-13}
        & LSTM & BERT & T5 & GPT-2 & Llama-2 &Anom. & LSTM & BERT & T5 & GPT-2 & Llama-2 &Anom. \\
        \midrule
        LIME & 58.47 & 55.78 & 66.66 & 51.46 & 54.35 & 62.30 & 0.604 & 0.584 & 0.603 & 0.533 & 0.521 & 0.575 \\
        LIME* & \textbf{75.09} &\textbf{68.20} & \textbf{86.99} & \textbf{69.03} & \textbf{62.68} & \textbf{80.10} & \textbf{0.887} & \textbf{0.927} & \textbf{0.924} & \textbf{0.759} & \textbf{0.728} & \textbf{0.763}\\[0.5em]
        KSHAP & 63.64 & 58.02 & 66.24 & 56.06 & 51.06 & 62.90 & 0.613 & 0.593 & 0.590 & 0.578 & 0.536 & 0.557\\
        KSHAP* & \textbf{86.10} & \textbf{83.75} & \textbf{73.62} & \textbf{69.34} & \textbf{61.81} & \textbf{77.40} & \textbf{0.879} & \textbf{0.890} & \textbf{0.909} & \textbf{0.711} & \textbf{0.680} & \textbf{0.716}\\[0.5em]
        DecompX & -- & 60.80  &--  & -- & -- & -- & -- & 0.601 & -- & -- &-- &-- \\
        \bottomrule
    \end{tabular}
    }
    \caption{Average accuracy and AUROC of the explanations generated by LIME, KSHAP, their \toolname-augmented versions, and DecompX for sentiment analysis models and the anomaly detection RNN (Anom.).
}
    \label{tab:fidelity-attr}
\end{table*}

\begin{figure*}[t]
    \centering
        \includegraphics[scale = 0.3]{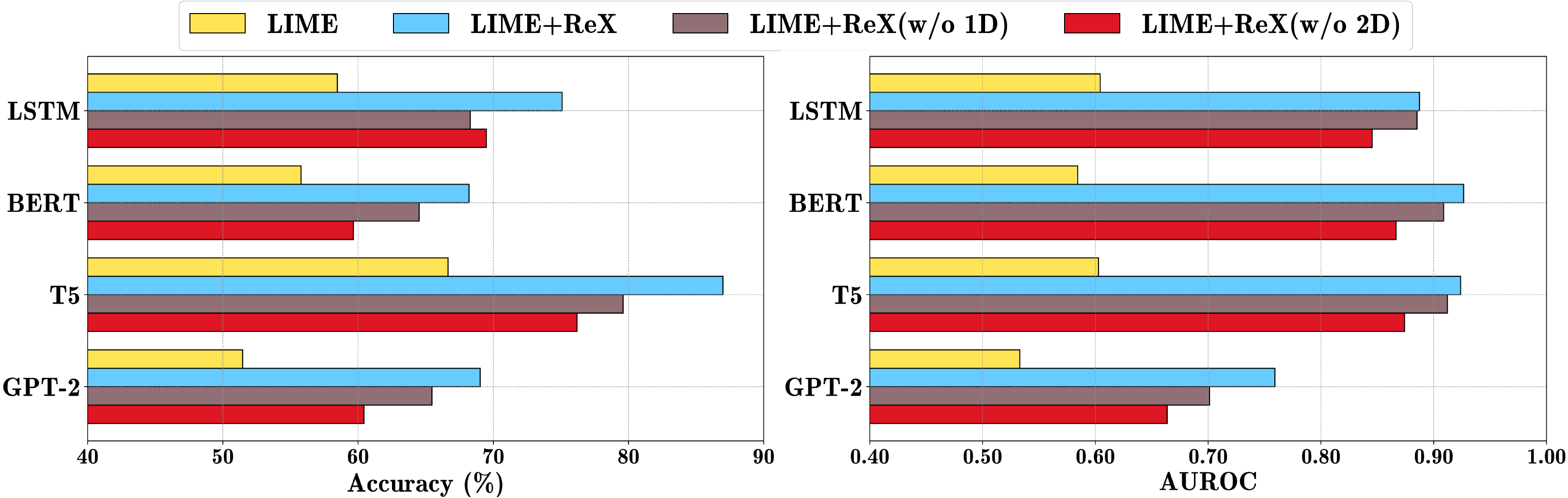} 
        
    
\caption{Average accuracy and AUROC of explanations for the four sentiment analysis models under different settings. The explanations are generated by LIME and its three augmented versions, which are augmented by \toolname, \toolname without 1D-predicates, and \toolname without 2D-predicates.
}
\label{fig:ablation}
\end{figure*}

In this section, we demonstrate the generality of \toolname and 
its effectiveness in improving the explanation fidelity and helping end users understand and predict the behaviors of the target models through empirical evaluation. 
To show the generality of \toolname, we instantiated it on three different techniques: Anchors, LIME, and Kernel SHAP (KSHAP for short). 
They were applied to explain various models of two classification tasks (sentiment analysis and anomaly detection), and a text-generation task.
To show the fidelity improvement by \toolname, we compared explanation fidelity of the \toolname-augmented techniques with the vanillas and a state-of-the-art model-specific technique, DecompX~\cite{DecompX}.
To show how much \toolname helps end users understand and predict the behaviors of the target model, we conducted a user study. Finally, we discuss the time efficiency of \toolname-augmented techniques.

\subsection{Target Models, Datasets and Experimental Setup}


\textbf{Sentiment Analysis.}
Sentiment analysis models take a text sequence as input and return a binary value indicating positive or negative sentiment.
We used an LSTM, BERT, GPT-2, and T5, along with Llama 2 as the target models,
and used the Stanford Sentiment Treebank dataset~\cite{socher2013recursive} with its original train/validation/test split. We explained the target models on the test set, which contains 1821 sentences.
For the text data, we set $\ran(x,i) =\{x_i\}$, and
defined $\per(x, i)$ as the set of words that BERT predicts can appear in the context of $x_i$.
In other words, all the vanilla feature predicates indicate whether the $i$-th word matches the $i$-th word in the target input and the vanilla perturbation model replaces words using BERT.
%
As long-distance temporal information exerts little influence on models, we set a window $w=5$, 
which further limits  $|d-k|\leq 5$ for $P_{1D}$, and $|d|\le 5$ and $l-k \leq 5$ for $P_{2D}$. 
%
%
%

\textbf{Anomaly Detection.}
Anomaly detection models take a real number sequence $x=(x_1,x_2,\dots, x_n)$ as the input. After reading $(x_1,x_2,\dots, x_i)$, the model outputs $y_i\in\{0,1\}$ that indicates if $x_i$ is an anomaly.
We trained an Anomaly Detection RNN~\cite{park2018anomaly} on an ECG dataset~\cite{UCRArchive2018} with its original train/validation/test split, 
and explained the target model on the 9 anomalous inputs in the test set.
For the real number sequence data, we set $\ran(x,i) =\{x_i\}$,
defined $\per(x, i)$ as the real numbers sampled from a normal distribution $\mathcal{N}(x_i, 1)$.
In other words, all the non-temporal features indicate whether the $i$-th number matches the $i$-th number in the target input, and the vanilla perturbation model samples numbers from a normal distribution for each feature.
For tractability, we limited the explanations to only consist of data points that are at most $20$ steps before the detected anomalous point in a time series.

\textbf{Text Generation.}
Text generation models take a text sequence as the input and return another text sequence.
Anchors, LIME, and KSHAP are not originally designed for text generation models, while MExGen~\cite{MExGen} introduces a method to adapt LIME and KSHAP to these models by converting the generation task to a regression task. We followed this approach to adapt LIME and KSHAP.
We also consider two additional baselines, C-LIME and L-SHAP, which are instances of MExGen that are designed specifically for text generation models.
We used LLama 2 as the target model, and explained it on 100 randomly chosen sentences from Google Natural Questions Dataset~\cite{qadata}. We used the same \toolname setting as the sentiment analysis task.



\subsection{Fidelity Evaluation}


%
Fidelity reflects how faithfully an explanation describes the target model.
As Anchors provides rule-based explanations while LIME and KSHAP provide attribution-based surrogates, we employ different metrics.

Considering that anchors are rule-based sufficient conditions~\cite{decisionsets,Anchors,decisionextracting1995}, we used \textbf{coverage} and \textbf{precision} as fidelity metrics.
Given a target model $f$, an input $x$, their corresponding anchor $g$, and the distribution $D$ from the perturbation model,
we define \textbf{coverage} as $cov(x;f, g) := \mathbb{E}_{z\sim D(x)}[g(z)]$, i.e., the proportion of inputs in the perturbation domain that \textbf{match the rules};
we define \textbf{precision} as $prec(x;f, g) := \mathbb{E}_{z\sim D(x)} [\displaystyle \mathbf{1}_\mathrm{f(x)=f(z)}|g(z)=1]$, i.e., the proportion of covered data that have \textbf{the same model output} as the original input.
%

For LIME and KSHAP, we use the metrics for surrogate models~\cite{balagopalan2022road,yeh2019fidelity,ismail2021improving}.
Given a target model $f$, an input $x$, their explanation surrogate model $g$, the distribution $D$, and a performance metric $L$~(e.g., accuracy, area under the receiver operating characteristic curve (AUROC), or mean squared error (MSE)), the (in)fidelity is defined as $\mathbb{E}_{z\sim D(x)}L (f(z), g(z))$. In our evaluation, we used accuracy and AUROC for the sentiment analysis task and anomaly detection task, and MSE for the text generation task.

%

Table~\ref{table:fidelity-anchor} shows the fidelity of anchors and \toolname-augmented anchors.
\toolname improves the average coverage by 218.7\% relative to the vanilla anchors,  
while maintaining roughly the same level of precision or slightly improving it.
%
%
Table~\ref{tab:fidelity-attr} and~\ref{tab:gene} show the explanation fidelity of LIME, KSHAP, and the \toolname-augmented versions.
On the sentiment analysis and anomaly detection task, relative to the vanillas, 
\toolname improves the average accuracy of explanations by {26.6\%} and  {26.3\%}, and the AUROC by  {45.8\%} and  {38.0\%} respectively;
on the text generation task, \toolname decreases the average MSE by 0.133 and 0.041 respectively, and both \toolname-augmented techniques outperform C-LIME and L-SHAP.
%
For these {20} setup pairs that are only different in whether \toolname is applied, our \textbf{paired t-tests} indicate that with over 99\% confidence, \toolname significantly improves the explanation fidelity. 

To illustrate the effect of 1-D and 2-D predicates, we conducted an ablation study on LIME. Figure~\ref{fig:ablation} shows the results, which indicate that both 1-D and 2-D temporal predicates can improve the fidelity separately, and bring about more significant improvement together.

We also compare the fidelity of \toolname with one of the state-of-the-art attribution-based techniques: 
DecompX~\cite{DecompX}, which is a white-box method designed for BERT. 
Since it is model-specific, it cannot be augmented with \toolname.
However, by applying \toolname, the explanation fidelity of LIME and KSHAP, two relatively old model-agnostic techniques, surpasses that of DecompX.
%


\begin{table}[t]
\centering
    {
    \setlength{\tabcolsep}{1mm}
    \small
\begin{tabular}{@{}lcccccc@{}} 
\toprule
Method & \text{LIME} & \text{C-LIME} & \text{LIME*}  & \text{KSHAP} & \text{L-SHAP}& \text{KSHAP*} \\
\midrule
MSE &0.187& 0.137 &\textbf{0.054}&0.069& 0.065 &\textbf{0.028}\\ 
\bottomrule
\end{tabular}
}
\caption{Average mean square error (MSE) of explanations for LLama 2 on the text generation task.}
\label{tab:gene}
\end{table}

\begin{table}[t]
    \centering
{
\setlength{\tabcolsep}{0.83mm}
    \small
\renewcommand{\arraystretch}{0.9}

\begin{tabular}{@{}lcccccccccc@{}}
\toprule
\multirow{2}{*}{Methods} & \multicolumn{5}{c}{\(\mathrm{Precision}_u (\%)\)} & \multicolumn{5}{c}{\(\mathrm{Coverage}_u (\%)\)} \\
\cmidrule(lr){2-6} \cmidrule(lr){7-11} 
& T1 & T2 & T3 & T4 & T5 & T1 & T2 & T3 & T4 & T5\\
\midrule
Anchors & 70.6 & 47.4 & 18.1 & 47.4 & 57.8 & 58.0 & 44.0 & 37.0 & 37.0 & 43.5 \\
Anchors* & \textbf{81.2} & \textbf{99.4} & \textbf{73.9} & \textbf{84.1} & \textbf{97.7} & \textbf{61.9} & \textbf{69.5} & \textbf{80.5} & \textbf{71.5} & \textbf{60.5}\\
\bottomrule
\end{tabular}
}
\caption{Average \(\mathrm{Precision}_u\) and \(\mathrm{Coverage}_u\) of each test in the user study.}
\label{table:user_study}
\end{table}

\begin{table*}[t]
    \centering
    {
        \small
    \setlength{\tabcolsep}{1mm}
\begin{tabular}{@{}lcccccccccccccc@{}}
    \toprule
        & \multicolumn{2}{c}{LSTM} & \multicolumn{2}{c}{BERT} & \multicolumn{2}{c}{T5} & \multicolumn{2}{c}{GPT-2} &\multicolumn{2}{c}{Llama 2 (senti.)}& \multicolumn{2}{c}{RNN (Anom.)} &\multicolumn{2}{c}{Llama 2 (gene.)}
        \\
        \cmidrule(lr){2-3}  \cmidrule(lr){4-5}
        \cmidrule(lr){6-7}  \cmidrule(lr){8-9}
        \cmidrule(lr){10-11}  \cmidrule(lr){12-13}
        \cmidrule (lr){14-15}
        & & * & & * & & * & & * & & * & &* & & *\\[-0.5em]
        \midrule
Anchor  &\textbf{0.43} & 0.76 & \textbf{1.87} & 1.92 & 11.06 & \textbf{10.03} & 14.05 & \textbf{8.72} & 612.78 & \textbf{276.54} & 460.30 & \textbf{371.40} & -- & --\\
LIME  &\textbf{0.15} & 3.98 & \textbf{1.23} & 5.04 & \textbf{2.82} & 7.41 & \textbf{4.12} & 8.27 & \textbf{239.55} & 241.76 & \textbf{245.20} & 263.30 & \textbf{398.77} & 404.08\\
KSHAP  &\textbf{0.13} & 3.97 & \textbf{1.23} & 5.45 & \textbf{3.50} & 7.93 & \textbf{4.02} & 8.08 & \textbf{289.38} & 297.90 & \textbf{267.40} & 291.40 & \textbf{424.55}& 429.61\\
    \bottomrule
\end{tabular}
    }
    \caption{Average execution time (in seconds) of Anchors, LIME, and KSHAP and their \toolname-augmented version (denoted as ``*'') to explain the models in our experiments.}
\label{table:time}
\end{table*}
\subsection{User Study}

To assess how \toolname helps end users understand target models and predict their behaviors, we conducted a user study by comparing anchors and \toolname-augmented anchors (denoted as anchors*) on the preceding sentiment analysis LSTM.
Similarly, we used coverage and precision as metrics, but now they describe how well a human's predictions match the model's after consuming explanations. These metrics are denoted as \(\mathrm{precision}_u\) and \(\mathrm{coverage}_u\).
%
We employed 19 CS undergraduates with machine learning backgrounds but no experience with explanation techniques, to study how much \toolname improves Anchors.
%
%
%
The questionnaire contains five tests.
Each test first presents a sentence, the network's output on the sentence, and their corresponding anchor and anchor*.
We randomly chose the sentences from the test set.
Then we asked each user to predict the RNN's output on 10 new sentences, which are produced using our perturbation model (with BERT).
They could answer ``positive'', ``negative'', or ``I don't know''.
\(\mathrm{Coverage}_u\) for a sentence is the percentage of users that do not answer ``I don't know''.
\(\mathrm{Precision}_u\) for a sentence is the percentage of users that give a prediction that matches the model output among the users that do not answer ``I don't know''.

%
Table~\ref{table:user_study} shows the average \(\mathrm{coverage}_u\) and \(\mathrm{precision}_u\) across the 19 users and 10 sentences for each test.
The anchors* outperform the anchors on all tests in terms of both \(\mathrm{coverage}_u\) and \(\mathrm{precision}_u\).
Among these tests, the anchors* yield an average \(\mathrm{precision}_u\) of 87.3\% and an average \(\mathrm{coverage}_u\) of 68.8\%, while the anchors yield only 48.3\% and 43.9\%.
The relative improvements are 80.9\% and 56.7\% respectively. 
We did \textbf{paired t-tests} on these paired data. With more than 99\% confidence, \toolname significantly helps users predict more instances more precisely, i.e., \toolname helps users better understand the target model's behavior.

\subsection{Runtime Overhead}
Table~\ref{table:time} shows the execution time of the fidelity experiments. 
For Anchors, \toolname increases the execution time of explaining the LSTM and BERT by 0.19 seconds on average, but reduces the time to explain other target models by 107.8 seconds on average;
for LIME and KSHAP, \toolname slightly increases the average execution time by 6.87 seconds.

How \toolname affects the explanation time depends on the underlying explanation technique. 
The execution time of Anchors heavily depends on the underlying KL-LUCB algorithm. \toolname can often accelerate the KL-LUCB algorithm.
The execution times of LIME and KSHAP equal the sum of the model's predicting time and the regression time. \toolname keeps the same predicting time and increases the regression time from $O(n^2|X_s|)$ to $O(n^2w^4|X_s|)$ in the sentiment analysis and text generation tasks, and from $O(20^2|X_s|)$ to $O(20^6|X_s|)$ in the anomaly detection task, where $n$ is the input length, $|X_s|$ is the number of samples, and $w=5$ is a small constant.
For small models, such an increase is acceptable as the original explanation techniques already run fast.
For large models like  LLama 2, the extra overhead is negligible as the explanation time is dominated by running the model. 

\section{Related Work}
\label{sec:related}
Our work is related to explanation techniques capturing temporal information and (model-agnostic or model-specific) local explanation techniques.

Within our knowledge, existing techniques that capture temporal information provide global explanations. These techniques mainly provide DFAs~\cite{Jacobsson05, WeissGY18, wang2023deepseer} and their variants~\cite{AyacheEG18,DuXLM0Z19,DongWSZWDDW20} as global surrogates.
However, as the complexity of practical target models increases, faithful global explanations are hard for users to understand, which limits these techniques to explaining relatively simple models.
%

In contrast, local explanation techniques generate explanations that are easier to understand, as they describe target models' behaviors on a subset of inputs. 
Existing local explanations describe model behaviors by presenting the effect of each input feature value, e.g., feature attribution~\cite{LIME, SHAP, StrumbeljK14,tan2023glime,ArrasMMS17,vinayavekhin2018focusing,ArrasOMS19, schlegel2019towards,TranSHAP,DenilDF14,MurdochLY18}, 
decision rules~\cite{Anchors,lore},
counterfactuals,~\cite{Counterfactual, DandlMBB20, ZhangSS18},
 or visualization~\cite{ICE,LiCHJ16, DingLLS17}.
For models processing variable-length inputs, such explanations cannot faithfully capture models' behavior. 
Therefore, a few model-specific techniques consider the effect of multiple features together~\cite{chen2020generating, singh2018hierarchical, sivill_limesegment_2022,tsang2020does,windowSHAP,mixing,valuezeroing},
but these techniques are designed for specific models or domains, and still ignore temporal information, thus limiting their fidelity.
\section{Limitations and Future Work}
\label{sec:limit}
Although we have demonstrated the effectiveness of \toolname, there are still some limitations remaining.

\textbf{Realistic Perturbation Models.} The perturbation model is a key component of model-agnostic explanation techniques. However, in some domains, finding a realistic perturbation model is challenging. 
\toolname also faces this challenge.
For example, the perturbation for time series data like stock prices is not clear.

\textbf{Efficiency.} \toolname increases the number of predicates. 
The benefits of temporal predicates reduce the running time of rule-based methods, but not for attribution-based methods. \toolname still slightly increases the running time of LIME and KSHAP. 
%
If we can eliminate unimportant predicates, we can further reduce the running time.
We plan to address this in our future work.
%

\section{Conclusion}
\label{sec:con}

In conclusion, we have proposed \toolname, a general framework that adds temporal information to existing local model-agnostic explanation techniques.
\toolname allows these methods to generate more useful explanations for models that handle inputs of variable lengths (e.g., RNNs and transformers).
\toolname achieves this by extending language components of explanations with temporal predicates, and modifying perturbation models so they can generate different-length samples.
We have instantiated \toolname on Anchors, LIME, and Kernel SHAP, and demonstrated the effectiveness empirically.

\section*{Acknowledgments}
This work was supported in part by the National Natural Science Foundation of China under Grant No. 62172017.

\small
\bibliography{rex}

\clearpage
\renewcommand{\thesection}{\Alph{section}}
\setcounter{section}{0}

\onecolumn

\section{Notations}
\begin{align}
    \centering
    \begin{tabular}{l|l}
        \hline
        Notations & Definition \\
        \hline
        $\llbracket a,b \rrbracket$ & The set of integers from $a$ to $b$. \\
        \hline
        $f:\mathbb{R}\rightarrow \mathbb{R}$ & A machine learning model. \\
        \hline
        $g$ & A local explanation, which describes the behavior of $f$ in a local region.\\
        \hline
        $\mathrm{per}$ & A perturbation domain.\\
        \hline
        $P$ & A set of predicates.\\
        \hline
        $p:\mathbb{R}^*\rightarrow \{0,1\}$ & A predicate.\\
        \hline
        $p_i$ & A feature predicate.\\
        \hline
        $p_{k,d,op}^{1D}$ & A 1D temporal predicate.\\
        \hline
        $p_{k,l,d,op}^{2D}$ & A 2D temporal predicate. \\
        \hline
        $P^R,t^R,\dots$ & The set of predicates, local-explanation methods, etc. augmented with \toolname.\\
        \hline
        $\mathrm{ran}$ & The range of a feature predicate.\\
        \hline
        $t$ & A local-explanation method.\\
        \hline
        $t_{per}$ & A perturbation model.\\
        \hline
        $w$ & The window size of temporal predicates.\\
        \hline
        $X_s$ & A set of samples generated by the perturbation model.\\
        \hline
        $x\in \mathbb{R}^{n}$ & An input to a machine learning model, usually the one to be explained.\\
        \hline
        $x_i,z_i \in \mathbb{R}$ & The $i$-th feature of input $x$ and $z$.\\
        \hline
        $z\in \mathbb{R}^{n}$ &  An input to a machine learning model, usually generated by perturbation models.\\
        \hline
        $\mathbb{R}$ & The set of real numbers. \\
        \hline
        $\mathbb{R}^*$ & The set of real number sequences of non-negative length. \\
        \hline
        $\mathbb{R}^{n}$ & The set of real number sequences of length $n$. \\
        \hline
        $\mathbb{Z}$ & The set of integers. \\
        \hline
    \end{tabular}
    \label{table:notations}
\end{align}


\section{The \toolname (tempoRal eXplanation) Framework (Continued)}

\label{app:ins}
In this section, we introduce the details of incorporating \toolname into existing local explanation methods. 
%

\subsection{Augmenting LIME} 

Given an input $x\in\mathbb{R}^n$, a machine learning model $f$, another input $z\in\mathbb{R}^n$ similar to $x$, and a set of feature predicates $P = \{p_{1}, p_{2}, \ldots, p_{n}\}$, a LIME explanation $g$ has the form of 
\[
    g(z) = \sum_{i=1}^{n} \omega_{i}p_{i}(z) + b,
    \]
where $\omega_i$ is the weight of the $i$-th feature predicate, and $b$ is the bias term. 
The linear model $g$ is trained to approximate the target model $f$ in the neighborhood of $x$ by minimizing the loss function $\mathcal{L}(f, g, \pi_{x}) + \Omega(g)$, where $\pi_{x}$ is a proximity measure and $\Omega(g)$ is a regularization term. 
Formally, a LIME explanation is generated by solving the following optimization problem:
\[
    \xi 
     = \arg\min_{g\in G} \mathcal{L}(f, g, \pi_{x}) + \Omega(g)
\]
where $G$ is the set of all possible explanations.

LIME generates explanations with the three steps we mentioned in Section~\ref{sec:pre}. We then introduce how to augment LIME with \toolname in each step.

\subsubsection{Producing Predicates}
LIME constructs the feature predicate set $P$ defined in equation~\ref{eq:p}.
\toolname extends the predicate set $P$ to $P^R$ with temporal predicates defined in equation~\ref{eq:1d} and~\ref{eq:2d}, i.e. $P^R = P \cup P_{1D} \cup P_{2D}$.

\subsubsection{Generating Samples}

\algrenewcommand\algorithmicrequire{\textbf{Input:}}
\algrenewcommand\algorithmicensure{\textbf{Output:}}
\algrenewcommand\algorithmiccomment[1]{\hfill \small \textit{\color{gray}// #1}}

\begin{algorithm}[t]
    \caption{$t_{per}$, Vanilla Perturbation Algorithm} 
    \begin{algorithmic}[1] 
        \Require $x$: an input to be explained, 
        \Statex \quad\  $n$: the number of samples
        \Ensure $z$: $n$ samples within the perturbation domain $D$
        \Statex \quad\quad\quad $pv$: the corresponding predicate vectors
        \State $z \gets \{x, x, \dots, x\}$ \Comment{Initialize the $n$ samples with $x$}
        \State $pv \gets \mathbf{1}_{n \times |P|}$ \Comment{Initialize the $n$ predicate vectors}
        \For{$i$ in $1 \to n$} \Comment{Iterate over each sample}
            \State $c \gets \text{random\_select}(\llbracket 1, |x|\rrbracket)$ \Comment{random\_select returns a subset of the given set}
            \For{$j$ in $c$}  \Comment{Change the selected feature values}
                \State $z[i,j] \gets$ random\_select$(\mathrm{per}(x,i))$ 
                 \State $pv[i,j] \gets 0$  \Comment{Update the predicate vector}
            \EndFor
        \EndFor
        \State \Return $z, pv$
    \end{algorithmic}
    \label{alg:limeper}
\end{algorithm}

\begin{algorithm}[t]
    \caption{$t_{per}^R$, \toolname-Augmented Perturbation Algorithm} 
    \begin{algorithmic}[1] 
        \Require $x$: an input to be explained, 
        \Statex \quad\  $n$: the number of samples
        \Ensure $z$: $n$ samples within the perturbation domain $D$
        \Statex \quad\quad\quad $pv$: the corresponding predicate vectors

        \State $pv \gets \mathbf{0}_{n \times |P^R|}$ \Comment{Initialize predicate vectors}

        \State $z,\_ \gets t_{per}(x,n)$ \Comment{Sampling with the vanilla algorithm}
        
        \For{$i \text{ in } 1 \to n$} \Comment{Randomly delete features}
            \State $deleted \gets \text{random\_select}([1, |x|])$ 
            \For{$j \in deleted$}
                \State $z[i,j] \gets$ \textbf{empty}
            \EndFor
        \EndFor

        \For{$i \text{ in } 1 \to n$} \Comment{Randomly swap features}
            \State $k, l \gets \text{random\_select}([1, |x|]^2)$
            \State swap($z[i,k], z[i,l]$)
        \EndFor

        \For{$i \text{ in } 1 \to n$} \Comment{Update the predicate vectors}
            \For{$j \text{ in } 1 \to |P^R|$}
                \State $pv[i,j] \gets P^R_j(z[i])$ \Comment{$P^R_j$ is the $j$-th predicate in $P^R$}
            \EndFor
        \EndFor
        \State \Return $z, pv$
    \end{algorithmic}
    \label{alg:rexper}
\end{algorithm}

LIME generates a set of sample $X_s$ by the perturbation model $t_{per}$. The perturbation algorithm is shown in Algorithm~\ref{alg:limeper}.

\toolname-augmented LIME use $t_{per}^R$ to generate variable length input as we described in Section~\ref{sec:frame}. The perturbation algorithm is shown in Algorithm~\ref{alg:rexper}.

\subsubsection{Learning Explanations}
After generating the samples, LIME also gets a predicate vector for each sample.
Specifically, for each sample $z$, its corresponding predicate vector $pv$ is a binary vector that indicates which predicates are satisfied by $z$. In other words, if the $i$-th predicate is satisfied by $z$, then $pv_i = 1$; otherwise, $pv_i = 0$.

For the vanilla LIME, an example matrix consisting of the predicate vectors and corresponding outputs is shown as follows:
\[
    \begin{tabular}{@{}c|c|c|c|c|c|c|c}
        \hline
        & $pv_1$ & $pv_2$ & $pv_3$ & $\cdots$ & $pv_{|x|-1}$ & $pv_{|x|}$ & $output$\\
        \hline
        $z_1$& 1 & 0 & 1 & $\cdots$ & 0 & 1 & 0.64\\
        $z_2$ &0 & 1 & 0 & $\cdots$ & 1 & 0 & 0.23\\
        $\vdots$ & $\vdots$ & $\vdots$ & $\vdots$ & $\ddots $ & $\vdots$ & $\vdots$ \\
        $z_n$& 1 & 1 & 0 & $\cdots$ & 1 & 0 & 0.98\\
        \hline
    \end{tabular}
\]
For \toolname-augmented LIME, an example the matrix is as follows:
\[
    \begin{tabular}{@{}c|c|c|c|c|c|c}
        \hline
        & $pv_1$ & $pv_2$  & $\cdots$ & $pv_{|P^R|-1}$ & $pv_{|P^R|}$ & $output$\\
        \hline
        $z_1$& 1 & 0 & $\cdots$ & 0 & 1 & 0.64\\
        $z_2$&0 & 1  & $\cdots$ & 1 & 0 & 0.23\\
        $\vdots$&$\vdots$  & $\vdots$ & $\ddots $ & $\vdots$ & $\vdots$ \\
        $z_n$&1 & 1 & $\cdots$ & 1 & 0 & 0.98\\
        \hline
    \end{tabular}
\]

With the predicate vectors and corresponding outputs, LIME and the \toolname-augmented version fit the local explanation $g$ with K-Lasso and Ridge Regression. 
The \toolname-augmented explanation is in the form of $g(z) = \sum_{p\in P}\ \omega_pp_(z) + b$, which contains temporal information by the temporal predicates in $P^R$.

\subsubsection{Augmenting Kernel SHAP} 
Kernel SHAP uses LIME as the base method and makes the weight of each feature predicate $\omega_i$ to be a Shapley value by changing $\Omega(g)$, $\pi_x$ and $L$ in the following equation:
\[
    \xi 
     = \arg\min_{g\in G} \mathcal{L}(f, g, \pi_{x}) + \Omega(g)
     \]
Specifically, Kernel SHAP sets
\[
\begin{array}{r@{}l}
    \Omega(g) &= 0,\\
    \pi_x(z) &= \frac{M-1}{M choose |z|}|z|(M-|z|),\\
    \mathcal{L}(f, g, \pi_{x})& = \sum_{z\in X_s} \pi_x(z)(f(z) - g(z))^2,
\end{array}
\]
where $|z|$ is the number of ones when convert $z$ into a binary vector by the predicates.

Since \toolname does not change the learning algorithm, so we can extend Kernel SHAP with similar steps to LIME.


\subsubsection{Augmenting Anchors} 
Anchors are sufficient conditions formed by a conjunct of feature predicates. 
The generating process uses a beam search algorithm with the KL-LUCB algorithm to iteratively add new predicates to the explanation until the explanation satisfies the user-defined precision. The KL-LUCB algorithm selects the next predicate to add to the explanation by sampling by $t_{per}$ and then uses the sample to update the confidence interval of the predicate.

Therefore, to augment Anchors with \toolname, we can directly replace the predicates set with $P^R$ and the perturbation model with $t_{per}^R$ as augmenting LIME.

\section{Experiment Details}

\subsection{Fidelity Experiment Settings}
\label{app:evelset}
We experimented on a machine with an i9-13900K CPU, 128 GiB RAM, and RTX 4090 GPU.

To measure the fidelity improvement brought by \toolname, we keep all hyperparameters the same for both vanilla and augmented methods.

For LIME and KSHAP, we set the number of sampled inputs to 10000 except for explaining Llama-2. We set it to 1000 since it takes a long time for Llama 2 to generate feedback. 

MExGen designed a segmentation algorithm for text generation input and modified the algorithm of LIME and SHAP to adapt them on text generation tasks. The adapted version is called C-LIME and L-SHAP.
For C-LIME and L-SHAP, since the inputs in the Google Natural Questions dataset are relatively shorter, we set the maximum length in segmentation to 3.

For Anchors and DecompX, we keep all the default settings.

In the sentiment analysis experiment, we use an LSTM with $embedding\_dim= 256, hidden\_dim = 128, num\_layers=2$, BERT (base), GPT-2 (medium), T5 (base), Llama2-7b-chat as target models.
In the anomaly detection experiment, we trained our RNN following~\cite{park2018anomaly}.
In the text generation task, we also use Llama2-7b-chat as a target model.

For the LLama 2 model, when applying it to the sentiment analysis task, we simply use the following prompt:
\begin{quote}
    From now on, you should act as a sentiment analysis neural network. You should classify the sentiment of a sentence into positive or negative. If the sentence is positive, you should reply 1. Otherwise, if it's negative, you should reply 0. There may be some words that are masked in the sentence, which are represented by \textless{}UNK\textgreater{}. The input sentence may be empty, which is represented by \textless{}EMPTY\textgreater{}. You will be given the sentences to be classified, and you should reply with the sentiment of the sentence by 1 or 0.\\
There are two examples:\\
Sentence:\\
I am good\\
Sentiment:\\
1\\
Sentence:\\
The movie is bad.\\
Sentiment:\\
0\\
You must follow this format. Then I'll give you the sentence. Remember Your reply should be only 1 or 0. Do not contain any other content in your response. The input sentence may be empty.\\
Sentence:\\
\{The given sentence\}\\
Sentiment:
\end{quote}

In other words, we just explain how LLama 2 itself classifies the sentiment of a sentence. When the probability of positive sentiment is needed, we use similar prompts to ask LLama 2 to reply with the probability of positive sentiment.
When applying it to the text generation task, we directly ask LLama 2 the question in the dataset and let it generate the answer.

\subsection{Additional Evaluation Results}
\label{appen:addeval}
\begin{figure*}[t]
    \centering
        \begin{tabular}{c}
        \ \includegraphics[scale = 0.66]{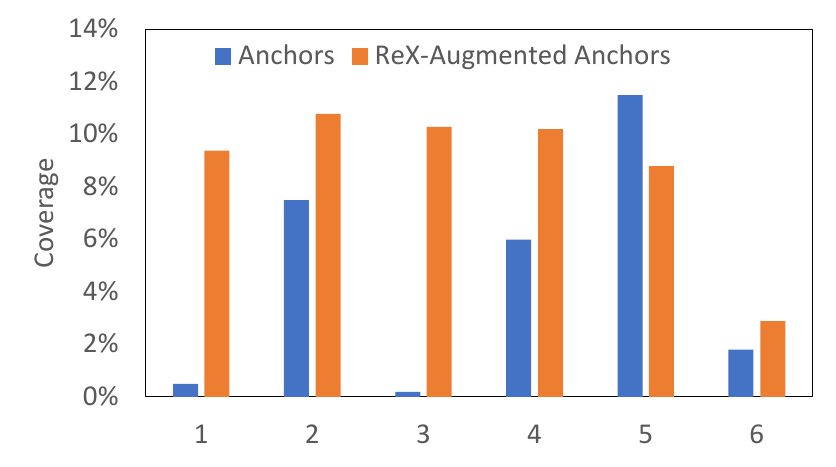}\\
    \end{tabular}
\caption{ Explanation coverage of all 6 inputs that the vanilla Anchors can generate an explanation under our setting in the anomaly detection experiment.}
\label{fig:covdetail}
\end{figure*}

\begin{table*}[ht]
\centering
\begin{tabular}{lcccccc}
\toprule
                   & LSTM            & BERT           & T5             & GPT-2          & Llama 2            & RNN(Anom.) \\ \midrule
LIME-accuracy      & 2.65e-8         & 2.53e-7        & 2.82e-8        & 3.71e-10       & 1.70e-141          & 2.16e-7        \\ 
LIME-AUROC         & 9.05e-28        & 4.41e-30       & 2.34e-28       & 1.21e-18       & 1.90e-41           & 1.62e-25       \\ 
KSHAP-accuracy     & 1.34e-10        & 5.90e-20       & 2.00e-6        & 1.01e-10       & 1.71e-9            & 3.06e-8        \\ 
KSHAP-AUROC        & 7.40e-30        & 1.39e-21       & 2.13e-30       & 1.05e-8        & 1.16e-10           & 1.82e-12       \\ 
Anchor-coverage    & 1.82e-7         & 2.39e-29       & 1.36e-37       & 1.17e-5        & 2.42e-7            & 2.66e-11       \\ \bottomrule
\end{tabular}
\caption{The p-value of each paired sample t-test on the sentiment analysis and anomaly detection tasks.}
\label{tab:ttest}
\end{table*}

Table~\ref{tab:ttest} shows the p-value of each Paired sample t-test in the sentiment analysis and anomaly detection task.
We do 6 t-tests for each statistic, and after Bonferroni correction, the p-values show that ReX significantly improves each statistic.
For the text generation task, the p-values of the LIME and KSHAP experiments are $6.42e-43$ and $4.65e-12$ respectively.

For explaining the anomaly detection model, we report the coverage of each explanation in Figure \ref{fig:covdetail} and omit the inputs that the vanilla Anchors cannot generate an explanation under our setting.
\toolname has improved the coverage for Anchors on most inputs.
On some inputs, the improvements are as high as 1,310.1\%.

\subsection{\toolname Parameter Settings}
\label{app:rexwindow}
Intuitively, using more predicates leads to longer execution time.
For example, in our sentiment analysis experiments, for an input with length $n$,
the vanilla techniques use $O(n)$ predicates, while the \toolname-augmented versions use $O(n^3)$ predicates by default. However, we can limit the number of introduced predicates according to the properties of specific tasks. For example, in our sentiment analysis experiments, we set $w=5$ to reduce the number of predicates to $O(nw^2)$, as long-distance temporal information exerts little influence on the model, 

Table~\ref{table:limeno-w},~\ref{table:anchorno-w}, and~\ref{table:shapno-w} shows the fidelity results of \toolname-augmented Anchors, LIME, KSAHP with and without setting $w=5$. They achieve similar fidelity, but the one with the $w$ limit is much faster. This is along with our intuition, so we set $w=5$ in our experiments.

\subsection{Computational Efficiency}
\label{sec:evaltime}
\begin{table}
    \centering
    {
        \setlength{\tabcolsep}{1.5mm}
\begin{tabular}{lccccc}
    \toprule
    & $w =$ & LSTM  & BERT  & T5  & GPT-2 \\
    \midrule
    \multirow{2}{*}{Accuracy (\%)} & $\infty$ & 73.93 & 69.45 & 87.74 & 68.66\\
    & 5 & 75.09 & 68.20 & 86.99 & 69.03\\[0.5em]
    \multirow{2}{*}{AUROC} & $\infty$ & 0.893 & 0.929 & 0.926 & 0.726\\
    & 5 & 0.887 & 0.927 & 0.924 & 0.759\\[0.5em]
    \multirow{2}{*}{Time (seconds)} & $\infty$ & 24.52 & 23.42 & 27.13 & 28.39\\
    & 5 & 3.98 & 5.04 & 7.41 & 8.27\\
    \bottomrule
\end{tabular}}
    \caption{Average accuracy, AUROC, and execution time of using \toolname-augmented LIME without and with setting $w=5$ to explain four sentiment analysis models.}
\label{table:limeno-w}

\end{table}

\begin{table}
    \centering
    {
        \setlength{\tabcolsep}{1.5mm}

\begin{tabular}{lccccc}
    \toprule
    & $w =$ & LSTM & BERT & T5 & GPT-2 \\
    \midrule
    \multirow{2}{*}{Precision (\%)} & $\infty$ & 85.27 & 83.36 & 85.67 & 82.62 \\
                                    & 5 & 86.48 & 84.04 & 85.04 & 82.54 \\
    \midrule
    \multirow{2}{*}{Coverage (\%)}  & $\infty$ & 9.64 & 10.77 & 11.53 & 9.15 \\
                                    & 5 & 10.77 & 12.07 & 11.78 & 10.26 \\
    \midrule
    \multirow{2}{*}{Time (seconds)} & $\infty$ & 12.96 & 18.14 & 37.4 & 40.4 \\
                                    & 5 & 0.76 & 1.92 & 10.0 & 8.72 \\
    \bottomrule
\end{tabular}}
    \caption{Average precision, coverage, and execution time of using \toolname-augmented Anchors without and with setting $w=5$ to explain four sentiment analysis models.}
\label{table:anchorno-w}
\end{table}

\begin{table}
    \centering
    {
        \setlength{\tabcolsep}{1.2mm}

    \begin{tabular}{lcccccc}
        \toprule
        & \multicolumn{2}{c}{$w = \infty$} & LSTM & BERT & T5 & GPT-2 \\
        \midrule
        \multirow{2}{*}{Accuracy (\%)} 
        & \multicolumn{2}{c}{$\infty$} & 85.03 & 81.06 & 69.70 & 69.99 \\
        & \multicolumn{2}{c}{5} & 86.10 & 83.75 & 73.62 & 69.34 \\
        \midrule
        \multirow{2}{*}{AUROC} 
        & \multicolumn{2}{c}{$\infty$} & 0.859 & 0.860 & 0.908 & 0.720 \\
        & \multicolumn{2}{c}{5} & 0.879 & 0.890 & 0.909 & 0.711 \\
        \midrule
        \multirow{2}{*}{Time (seconds)} 
        & \multicolumn{2}{c}{$\infty$} & 70.10 & 72.23 & 85.68 & 84.74 \\
        & \multicolumn{2}{c}{5} & 3.97 & 5.45 & 7.93 & 8.08 \\
        \bottomrule
    \end{tabular}
    }
    \caption{Average accuracy, AUROC, and execution time of using \toolname-augmented SHAP without and with setting $w=5$ to explain four sentiment analysis models.}
    \label{table:shapno-w}
\end{table}


%

%


%

The execution time of LIME and KSHAP is the sum of the target model and the regression algorithm.
Since \toolname does not increase the number of samples, the former one is the same.
Let n denote the length of the input $x$.
In vanilla techniques, the time of regression algorithm is $O(|P|^2|X_s|)=O(n^2|X_s|)$, while the time of \toolname-augmented techniques is $O(|P^R|^2|X_s|) = O(n^2w^4|X_s|)$.
This indicates that \toolname-augmented techniques spend more time on regression.
However, the regression algorithm is not the bottleneck. Table \ref{table:time} shows that the execution time of the \toolname-augmented techniques is about 4s slower than the vanillas. When explaining a large target model like Llama 2, such a difference is negligible.

The execution time of Anchors heavily depends on how quickly the underlying KL-LUCB algorithm~\cite{KL-LUCB} can find predicates that achieve the precision threshold.
KL-LUCB is a heuristic algorithm that iteratively finds a predicate to add to the explanation. 
Two factors affect the execution time: the number of iterations and the time to find a predicate in each iteration, corresponding to the quality of the predicates and the number of predicates.
Our experiments show that the execution times of Anchors and \toolname-augmented Anchors are comparable, and sometimes \toolname even accelerates Anchors.

In the anomaly detection experiments, we make the predicate set $P$ contain $20$ predicates, and $P_{1D}$ and $P_{2D}$ contain $20^3$ predicates. The situation is similar to the sentiment analysis experiments.

In summary, for attribution-based techniques like LIME and KSHAP, \toolname slightly increases the execution time; for rule-based techniques like Anchors, \toolname does not slow them, but sometimes reduces the execution time.

\textbf{In our sentiment analysis experiments, for an input with length $n$,}
the vanilla predicate set $P$ contains $O(n)$ predicates.
Since we set $w=5$, the predicate set $P_{1D}$ contains $O(n\times w)$ predicates, and $P_{2D}$ contains $O(n\times w^2)$ predicates.
So \toolname-augmented methods use $O(n\times w^2)$ predicates. Without the $w$, \toolname-augmented methods use $O(n^3)$ predicates.
Table~\ref{table:limeno-w} shows the results of using LIME* with and without $w$ limit. They achieve similar fidelity, but the one with the $w$ limit is much faster. This is along with our intuition, so we set $w=5$ in our experiments.

%
For LIME and KSHAP, the running time is the running time of the target model plus the regression algorithm.
Since \toolname requires the same number of sampled instances, the former one is the same.
In vanilla methods, the regression algorithm takes $O(|P|^2|X_s|)=O(n^2|X_s|)$ time, while \toolname-augmented methods take $O(|P^R|^2|X_s|) = O(n^2w^4|X_s|)$ time.
So \toolname-augmented methods are slower than vanilla method. However, the regression algorithm is not the bottleneck. Table \ref{table:time} shows that the running time of \toolname-augmented methods is about 4s slower than vanilla methods. When explaining a large target model like Llama-2, such a difference is negligible.

Anchors' runtime heavily depends on how quickly the underlying KL-LUCB algorithm~\citep{KL-LUCB} can find predicates that achieve the precision threshold.
KL-LUCB is a heuristic algorithm that iteratively finds a predicate to add to the explanation. 
Two factors affect the running time: the number of iterations and the time to find a predicate in each iteration, corresponding to the quality of the predicates and the number of predicates.
Our experiments show that the running times of Anchors and Anchors* are comparable, and sometimes \toolname even accelerates Anchors.

In the anomaly detection experiments, we make the predicate set $P$ contain $20$ predicates, and $P_{1D}$ and $P_{2D}$ contain $20^3$ predicates. The situation is similar to the sentiment analysis experiments.

In summary, for attribution-based methods like LIME and KSHAP, \toolname slightly increases the running time; for rule-based methods like Anchors, \toolname does not slow them and sometimes reduces the running time.
%
%
%

\subsection{User Study Details}
\begin{figure*}[t]
\centering
    \begin{tabular}{rl}
         \textbf{Original sentence}: & pretentious editing ruins a potentially terrific flick.  \\
         \textbf{RNN output:} & negative. \\
         \textbf{Explanation 1}:& the word ``ruins'' appears in the sentence at the specific position. \\
         \textbf{Explanation 2}: & both ``ruins" and ``terrific"  appear in the sentence,
 ``terrific" is behind ``ruins",  \\
 & and there are at least ``0" words between them.
    \end{tabular}
    \\[8pt]
    Please predict the RNN output on each sentence below according to each explanation. You can answer 0. negative, 1. positive, or 2. I don't know.
    \\[8pt]
    \begin{tabular}{l|l|l}
    \hline
         \textbf{Sentence}& \textbf{Prediction 1} & \textbf{Prediction 2} \\
         \hline
         ruins beneath terrific design. & & \\
         pretentious editing ruins a potentially terrific methodology. & & \\
         cult ruins a potentially lucrative planet. & &\\
         pretentious editing ruins describe potentially difficult behavioral. & & \\
 editing ruins potentially terrific circumstances. & &\\
 blue ruins are terrific. & &\\
 the on ruins a wonderful but flick theater. & & \\
 pretentious editing ruins potentially difficult solution. & & \\
 ice ruins potentially terrific. & &\\
 editing ruins without terrific positions. & &\\
         \hline
    \end{tabular}
    \caption{One of five tests in the user study.}
    \label{fig:study_example}
\end{figure*}
\label{appen:userstudy}
We introduce more details about the user study in this section.
\subsubsection{Additional Settings}

The questionnaires are similar for all users with minor variations in the order of presentation. We presented the five questions, Q1, Q2,…, and Q5 in a random order. For each question, we presented the explanations generated without/with ReX in a random order. Also, we presented the ten sentences to be predicted in a random order.
Figure~\ref{fig:study_example} shows one test question in the questionnaire.

\subsubsection{User Screening}
We recruited the users from students who selected a machine learning course at our university, and screened them before filling out the questionnaires, so the set is small.

To make sure our users understood the meaning of the explanation and would predict the model output based on the explanation rather than their own feelings about a text sentence, we make up a sentence with a predicted sentiment different from its sentiment in the real world, and present a simple if-then rule as an explanation. Then we let our users predict the sentiment of another ten sentences. We remain the users that strictly follow the if-then rule in our user study.

\subsubsection{Additional Results}

Among all the 50 sentences to be predicted, in 19 sentences, the prediction by the LSTM model differs from ground truth (the sentiment classified by humans). In these sentences, ReX can still help people to predict better. Users make the same prediction as the LSTM model on 48.9\% of these instances with anchor explanations, while ReX improves this to 66.7\%.
In other words, ReX helps humans make the same “wrong predictions” made by the model.

\subsubsection{Other information}
\label{appen:userirb}
The participants of our user study will not incur potential risk, and we have told them the description and instructions of our study before they join. Our study is approved by the IRB or our school.

\section{Broader Impacts}
\label{app:broader}
This paper proposed \toolname, which extends existing local model-agnostic techniques to capture temporal information, which plays an important role in ML models that process input of variable length.
The contribution of this paper is fundamental and will have a broader impact on a vast number of applications, 
especially since many transformer models and LLMs have been recently been depoyed in various domain fields.
\toolname can help end users understand these models' decisions and increase their trust in these models.
\toolname also brings benefits to the application of these models in terms of explainability, transparency, and fairness.
Because we only provide relatively simple local explanations, users may use them for neural network attacks or develop other negative impressions.

\end{document}